\documentclass[letterpaper]{article} 
\usepackage{aaai25}  
\usepackage{times}  
\usepackage{helvet}  
\usepackage{courier}  
\usepackage[hyphens]{url}  
\usepackage{graphicx} 
\usepackage{natbib}  
\usepackage{caption} 
\usepackage{algorithm}
\usepackage{algorithmic}
\usepackage{booktabs}

\usepackage{newfloat}
\usepackage{listings}
\DeclareCaptionStyle{ruled}{labelfont=normalfont,labelsep=colon,strut=off} 
\floatstyle{ruled}
\newfloat{listing}{tb}{lst}{}
\floatname{listing}{Listing}
\title{S-INF: Towards Realistic Indoor Scene Synthesis via Scene Implicit Neural Field}
\author{
    Zixi Liang\textsuperscript{\rm 1}, 
    Guowei Xu\textsuperscript{\rm 1}, 
    Haifeng Wu\textsuperscript{\rm 2}, 
    Ye Huang\textsuperscript{\rm 1}, 
    Wen Li\textsuperscript{\rm 1,\rm 2}\thanks{Corresponding author.}, 
    Lixin Duan\textsuperscript{\rm 1,\rm 3}
}
\affiliations{
    \textsuperscript{\rm 1}Shenzhen Institute for Advanced Study, University of Electronic Science and Technology of China\\
    \textsuperscript{\rm 2}School of Computer Science and Engineering, University of Electronic Science and Technology of China\\
    \textsuperscript{\rm 3}Sichuan Provincial Key Laboratory for Human Disease Gene Study and the Center for Medical Genetics,  Department of Laboratory Medicine, Sichuan Academy of Medical Sciences and Sichuan Provincial People's Hospital, UESTC\\

    \{zixiliang.acad, xuguowei368, haifengwu205, liwenbnu, lxduan\}@gmail.com, edward.ye.huang@qq.com
}

\usepackage{bibentry}

\begin{document}

\maketitle

\begin{abstract}

Learning-based methods have become increasingly popular in 3D indoor scene synthesis (ISS), showing superior performance over traditional optimization-based approaches.
These learning-based methods typically model distributions on simple yet explicit scene representations using generative models.
However, due to the oversimplified explicit representations that overlook detailed information and the lack of guidance from multimodal relationships within the scene, most learning-based methods struggle to generate indoor scenes with realistic object arrangements and styles.
In this paper, we introduce a new method, Scene Implicit Neural Field (S-INF), for indoor scene synthesis, aiming to learn meaningful representations of multimodal relationships, to enhance the realism of indoor scene synthesis.
S-INF assumes that the scene layout is often related to the object-detailed information. It disentangles the multimodal relationships into scene layout relationships and detailed object relationships, fusing them later through implicit neural fields (INFs).
By learning specialized scene layout relationships and projecting them into S-INF, we achieve a realistic generation of scene layout.
Additionally, S-INF captures dense and detailed object relationships through differentiable rendering, ensuring stylistic consistency across objects.
Through extensive experiments on the benchmark 3D-FRONT dataset, we demonstrate that our method consistently achieves state-of-the-art performance under different types of ISS.
\end{abstract}

%
\begin{links}
    \link{Code}{https://github.com/ZixiLiang/S-INF}
\end{links}

\section{Introduction}

\begin{figure}[!t]
  \setlength{\leftskip}{-4mm}
  \includegraphics[width = 0.52\textwidth]{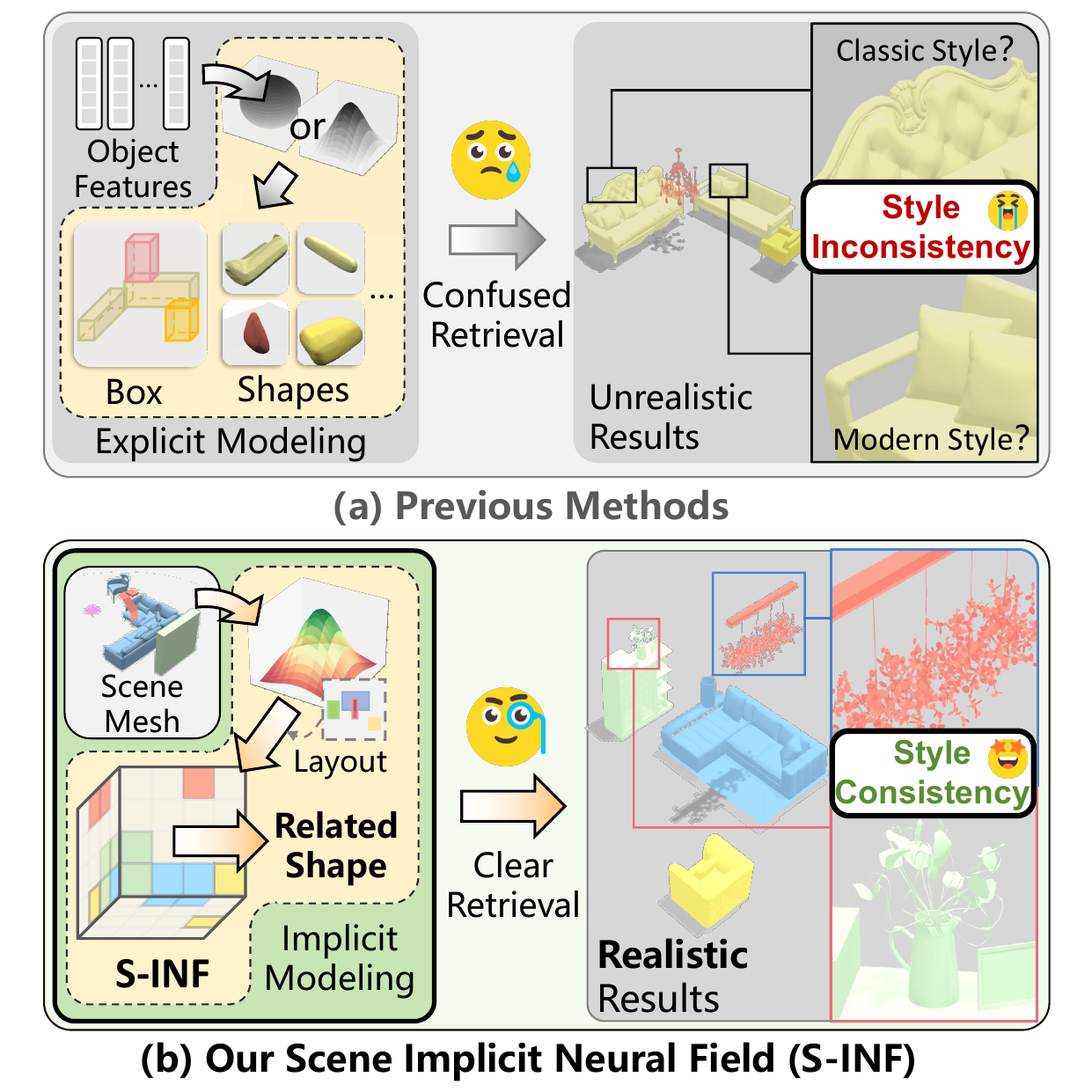}
  \caption{
  Unlike previous methods, we enhance the implicit modeling process from the S-INF with scene layout relationships and detailed object relationships, to achieve more layout realistic and style consistency generations.
  }
  \label{fig:intro}
\end{figure}

Synthesizing realistic and diverse 3D indoor scenes is a long-standing problem in computer vision and graphics.
This research topic has received widespread attention due to its significant cost reduction in fields such as virtual reality~\cite{vr1,vr2} and 3D design~\cite{design1,design2}.
Specifically, ISS can virtually rearrange existing furniture, enabling more convenient virtual interior design.
Despite recent progress on this topic~\cite{scenehgn,  sync2gen, layoutdm, roomdesigner, commonscenes}, the nature of the underlying multimodal distribution, including scene layout and detailed object relationships, makes it still challenging to generate realistic and diverse 3D indoor scenes.

At the outset of this research, scene modeling and synthesis were typically formulated as an optimization problem~\cite{survey}.
Using scene priors like room design rules~\cite{graph1,graph2} and in a human-centric manner~\cite{human1, human2, human3}, they first sample an initial scene and then refine its configuration through iterative optimization.
However, defining precise rules demands significant expertise and may limit the representation of complex, diverse scenes.
In recent years, learning-based methods have become popular, utilizing generative models to learn scene distributions from data, such as Generative Adversarial Networks (GANs)~\cite{gan1,gan2}, Variational Autoencoders (VAEs)~\cite{sync2gen,sg-vae,vae2}, and diffusion models~\cite{diffusion1,diffuscene,echoscene}.
These methods use generative models to model distributions on over-simplified and explicit-format scene representations (e.g., boxes and features).
They usually map these explicit scene representations to latent distributions, constrained by prior distributions such as Gaussian or spherical distributions.
When generating scenes, these methods decode sampled vectors from the prior distribution to obtain scene representations, followed by post-processing steps~\cite{context, structure, automatic, uni3d} to retrieve CAD models from the dataset and produce the final results.
While this approach establishes a solid generative framework for ISS tasks, the overly simplified explicit representations overlook scene layout relationships and lack guidance of detailed object relationships within the scene, hindering the model from effectively learning multimodal scene relationships.
Consequently, these methods often struggle to generate realistic indoor scenes, as they tend to focus on the major modes of the latent distribution while usually ignoring minor modes, a phenomenon known as mode collapse.
Specifically, due to the limited expressiveness of the learned latent distribution, it is challenging for them to generate complex scenes that have realistic relationships or stylistic consistency.

In this paper, we aim to address the limitations mentioned above by proposing the S-INF to model multimodal relationships.
S-INF assumes that the scene layout is often related to the object-detailed information.
It disentangles the multimodal relationships into scene layout relationships and detailed object relationships.
To generate indoor scenes, we first decode the scene layout relationships and detailed object relationships into the layout and the INF, then project the layout into the INF to obtain refined related shapes for retrieval.
The disentangling construction offers several advantages: 1) we directly extract more advantageous multimodal information from the entire scene in a multiscale manner and map them into the S-INF, effectively modeling the multimodal relationship within the scene.
2) Unlike previous methods that sample directly from the prior distribution and decode it into explicit scene representations, our latent space also learns detailed object relationships, leading to a style-consistancy generation.
3) During the learning process of the S-INF, we use differentiable rendering to capture dense and detailed object relationships, thereby ensuring realistic and stylistic consistency across objects (see Figure \ref{fig:intro}).
Based on this, we retrieve CAD models from the dataset according to refined meshes sampled from the S-INF to obtain the final result. This multimodal relationship-based related shapes helps achieve diverse and realistic indoor scene synthesis.
In summary, our main contributions are as follows:
\begin{itemize}
\item We uncover that the overly simplified explicit representations in current scene generation frameworks overlook detailed information and lack the necessary guidance for latent scene space modeling, making it difficult to effectively learn meaningful multimodal relationships, which leads to challenges in generating realistic indoor scenes.
\item We introduce a novel approach called Scene Implicit Neural Field (S-INF), which models wide and scene layout relationships as well as detailed object relationships, resulting in more realistic and style consistancy ISS.
\item Through extensive experiments conducted on the 3D-FRONT dataset, we demonstrate that our method attains state-of-the-art performance in ISS.
\end{itemize}

\section{Related Work}

\subsection{3D Indoor Scene Synthesis}

Early research treated the ISS task as an optimization problem, promoting the optimization process by introducing scene priors\cite{survey}. 
These priors typically included interior design guidelines, object frequency distributions, and scene arrangement examples. 
Guided by scene priors, new scenes can be generated from the formulation using various optimization methods, such as iterative approaches~\cite{iterapp1,iterapp2}, nonlinear optimization~\cite{nonlinear1,nonlinear2,nonlinear3}, or manual interaction~\cite{human1,human2,human3}.

Recently, many learning-based methods have been proposed to synthesize complex scene compositions.
These methods process the scenes to obtain simple, explicit scene representations (e.g., boxes and shapes), assuming that the scene representations obey a latent distribution.
Generative models, such as feed-forward networks~\cite{pose2room}, recurrent networks~\cite{grains,atiss,deep,planit,sceneformer,layoutdm,mime}, GANs~\cite{gan1,gan2}, VAEs~\cite{sync2gen,sg-vae,vae2}, and diffusion models~\cite{diffusion1,diffuscene,echoscene, instructscene}, are then designed to learn this latent distribution, followed by retrieving CAD models from datasets based on the output of the generative models, such as shape-based retrieval or box-based retrieval.
While the choice of generative models for ISS is crucial, the specific properties of the latent distribution that are advantageous for ISS remain unclear.
Many works incorporated Gaussian or spherical distributions into latent representations and focused on developing robust generative models.
However, directly modeling latent distributions based on explicit scene representations and lacking appropriate guidance on the latent distribution, these methods struggle to learn meaningful object relationship representations within scenes, thus achieving diverse and realistic 3D indoor scene synthesis.
Therefore, the goal of this paper is to achieve meaningful object relationship representations in the latent space, enabling the model to autonomously learn the local relationships between different objects and thereby achieve accurate object relationships.

\begin{figure*}[!ht]
  \centering
  \includegraphics[width = \textwidth]{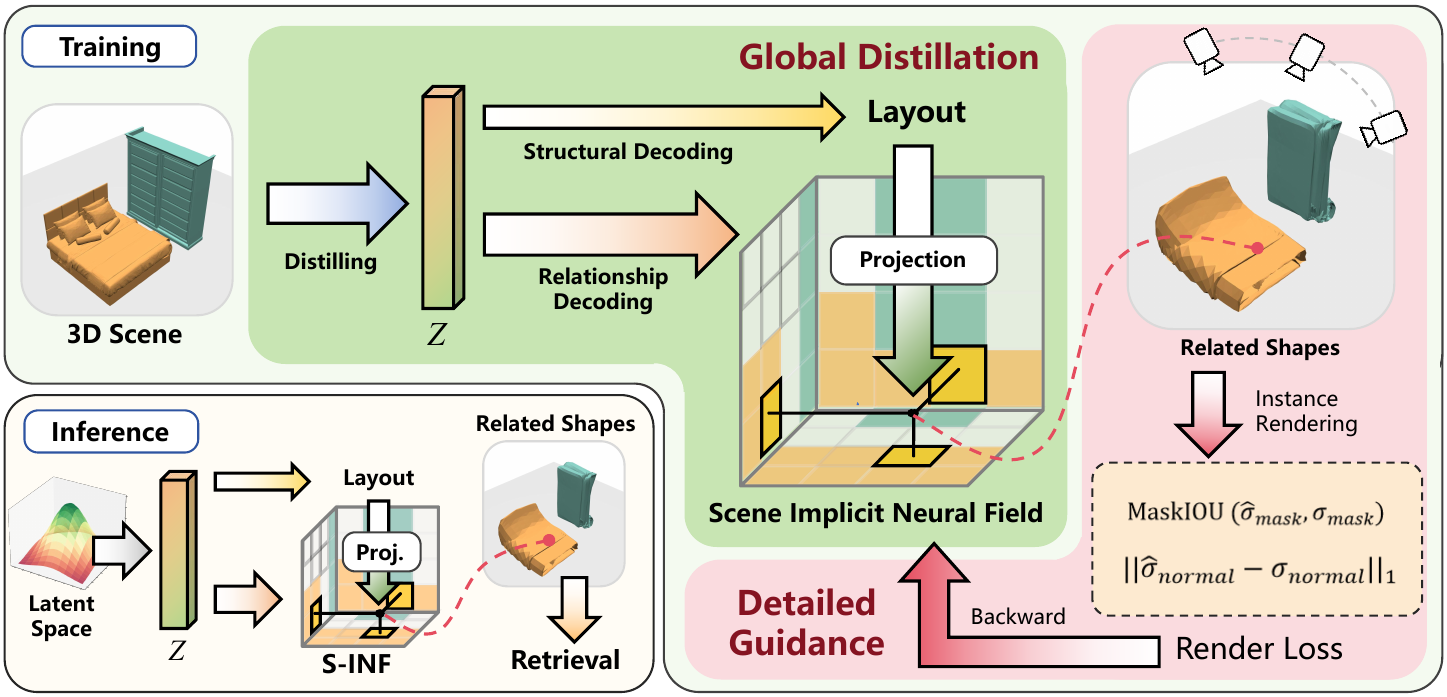}
  \caption{Our approach focuses on developing the S-INF to enable efficient capture of multimodal relationships and generate realistic and reliable 3D indoor scenes. We utilize the scene encoder to distill the realistic multimodal relationships into the S-INF. We also use differentiable rendering to enhance the S-INF style consistency information in detailed style relationships. Those optimized equip the S-INF with genuine multimodal relationship understanding capabilities, facilitating the generation of realistic,  and style-invariant 3D indoor scenes.}
  \label{fig:method}
\end{figure*}

\subsection{Implicit Neural Field}
Our S-INF is inspired by INFs, which have shown promising results in representing high-fidelity geometry and appearance in 3D.
Early work like DeepSDF~\cite{deepsdf} represents the shape of a class of objects using signed distance functions, allowing for high-quality interpolation and completion from partial and noisy 3D input data.
Neural Radiance Fields (NeRF)~\cite{nerf} leverage an MLP to model a coordinate-based radiance field, generating photo-realistic 2D renderings from novel views through volumetric rendering.
DualOGNN~\cite{dualOCNN} employs octrees to store the volumetric field of 3D shapes, effectively capturing shape details and demonstrating superior performance in various 3D shape and scene reconstruction tasks.
NKField~\cite{nkf,nksr} uses data-dependent neural kernels to encode INFs, demonstrating strong generalization capabilities in 3D scene completion tasks.
However, INF have not yet been fully explored in ISS.
This paper constructs an INF in the latent space to achieve meaningful multimodal relationship representations. This enables the model to autonomously learn the detailed object relationships between different objects and thereby achieve accurate object relationships.

\section{Methodology}
The objective of ISS is to generate a sequence of object meshes, $\hat{X} = \{\hat{x}_j\}_{j=1}^m$, where $m$ denotes the number of objects. 
The objects $\{\hat{x}_j\}_{j=1}^m$ should maintain object detailed consistency and a reasonable layout, and the modeled distribution $p(\hat{X})$ should closely match the diversity GT distribution $p(X)$.
In this section, we first review the general formulation of the Learning-based ISS. 
Based on the general framework, we then introduce our core S-INF equipped with global layout distillation and detailed style guidance. 
Finally, we summarize the training and inference processes.

\subsection{General Formulation of Learning-based ISS}

The core problem of learning-based ISS is to model the distribution $p(X)$.
It is nontrivial to directly parameterize this distribution with a neural network. 
To this end, existing methods usually assume the scenes can be decoupled into scene-wise layout information and object-wise detailed information, modeling the distribution $p(X)$ through a two-step sequential process.
The distribution $p(X)$ can be re-formulated as $p(X) = p(\{D_{IS}(b_i)\}_{i=1}^n | b_i \in D_{RS}(z) \}_{i=1}^n)$, where the Relationship Decoder $D_{s}$ captures scene relationships, the Instance Decoder $D_{IS}$ learns object representations, and $z$ is noise sampled from a prior distribution, such as Gaussian or Spherical distribution.
The formulation of the above process can be summarized as follows:
\begin{equation}\label{eq:previous_scene}
    \{b_i\}_{i=1}^n = D_{RS}(z), \ \ z\sim p(z),
\end{equation}
\begin{equation}\label{eq:previous_obj}
    \{x_i\}_{i=1}^n = \{D_{IS}(b_i)\}_{i=1}^n.
\end{equation}

\noindent The incorporation of the scene layout representation $\{b_i\}_{i=1}^n$ streamline the scene generation learning process. 
While generative models theoretically have the potential to incorporate a wide variety of scene attributes for $z$ to learn, the discrete modeling of explicit object representation hinders information transfer between objects, making it difficult to accurately capture multimodal relationships, including layout-modal and object-modal.
Specifically, the $\{b_i\}_{i=1}^n$ in Eq.~(\ref{eq:previous_scene}) are often with the over-simplified explicit format, and discretized, lacking expressiveness in the details of objects.
The independent decoding of objects in Eq.~(\ref{eq:previous_obj}) further leads to inconsistencies in object-detailed style due to the lack of constraints on detailed object relationships~\cite{sceneprior}.

\begin{figure*}[!ht]
\centering
\includegraphics[width = \textwidth]{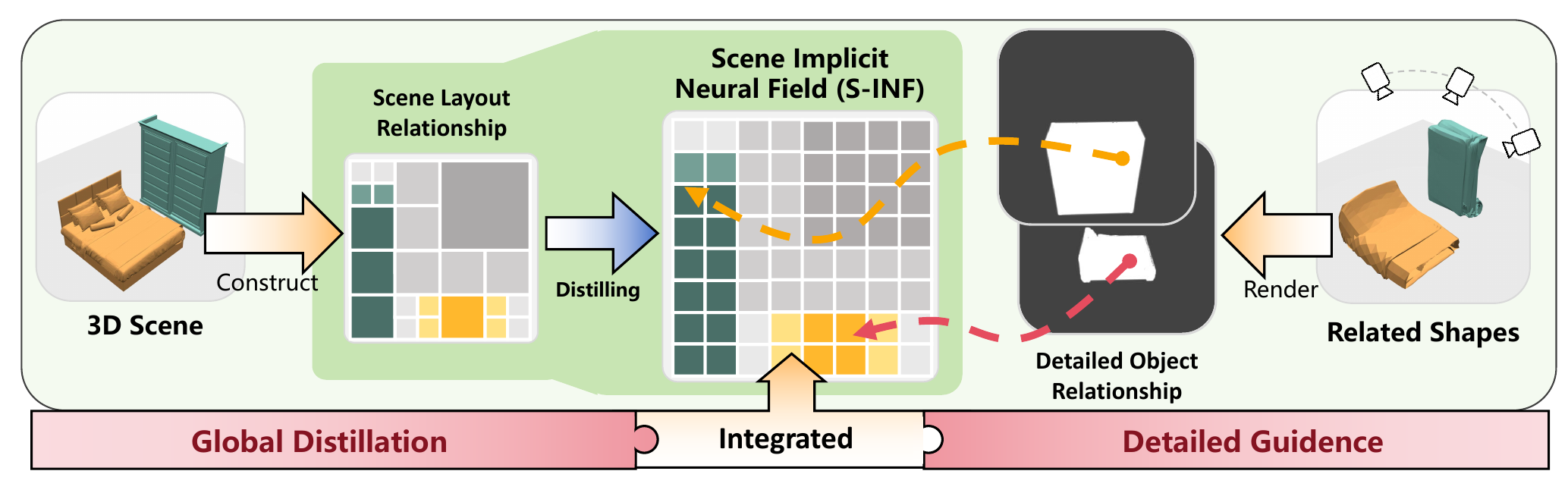}
\caption{
Integration between scene layout relationships and detailed object relationships.
We leverage a specialized scene encoder to construct scene layout relationships (and little detailed object relationships) and distill multimodal relationships into S-INF. On the other hand, the rendered image provides dense and detailed style information, enriching the detailed object relationships in S-INF.
}
\label{fig:field}
\end{figure*}

\subsection{Scene Implicit Neural Field}
To address the above issues, we construct the S-INF in the latent scene space, benefiting from its implicit modeling capabilities for potentially advantageous features, which is depicted in Fig.~\ref{fig:method}.
The latent vector is decomposed into two factors: the scene layout representation $l_{b}$ and the INF $f_b$. 
The scene representation, enriched with multimodal relationships, is assumed to be generated by sampling the scene layout within the INF. This process is formalized as follows:
\begin{equation}\label{eq:our_scene}
    f_b,l_{b} = D_{RS}(z), \ \ z\sim p(z),
\end{equation}
\begin{equation}\label{eq:our_obj}
    \{x_i\}_{i=1}^n = ( D_{IS} \circ f_b)(l_b),
\end{equation}
where operation $g\circ  h: X\to  Z$ denotes the function composition of $g: Y\to Z$ and $h: X\to Y$.
In implementation, we draw from current works~\cite{get3d,tri-plane1,tri-plane2} and employ a 2D CNN to map the latent vector $z$ to the INF of dimensions $N \times N \times (C \times 3)$. 
Here, $N$ denotes the spatial resolution, and $C$ represents the feature channels in the field, the output vector $f_b(p)$ of a point $p$ can be interpolated as $f_b(p) = {\textstyle \sum_{e}\rho [\pi_e(p)]}$, where $\pi_e$ denotes the projection of the point $p$ onto feature plane $e$, and $\rho$ represents the bilinear interpolation operation. 
We employ a Transformer~\cite{transformer} decoder and generate the transformation of template spheres for deformation.

As shown in Eq.~(\ref{eq:our_scene}), S-INF disentangles multimodal relationships into $f_b$, which captures detailed object relationships, and $l_b$, which captures scene layout relationships.
Through such explicit disentangling, S-INF demonstrates a superior capability in representing multimodal relationships compared to simply using Tri-Plane INF.
In addition, by integrating global distillation and detailed guidance, S-INF can provide realistic multimodal relationships and style-consistent related shapes $\{x_i\}_{i=1}^n$ for retrieval.
The following sections will introduce these two approaches.

\subsubsection{Global Distillation.}\label{sec:ocnn}
We utilize the global distillation to extra multimodal relationships globally from the Scene Encoder $E_{S}$, storing the extracted scene layout relationships and detailed object relationships into the layout $l_b$ and INF $f_b$ of S-INF, respectively. 
These are then combined into the related shapes $\{x_i\}_{i=1}^n$ through the process described in Eq.~(\ref{eq:our_obj}).
Current methods typically rely on manually defining the over-simple explicit object representations, such as boxes, to represent the scene layout, then learning the scene layout relationships from it.
However, we contend that these overly simplistic explicit representations lose detailed object relationships, leading to unrealistic layouts, such as overlaps, error arrangements, and misalignments (see Fig.~\ref{fig:result}). 
Additionally, these approaches make it difficult for the model to learn complex scene layout relationships, such as embedded relationships.

Fortunately, the implicit modeling capability of INFs allows for the accommodation of detailed object structural information and facilitates learning of complex scene layout relationships~\cite{proof_field,tri-plane1}. This enables us to combine INFs with explicit layouts to generate more realistic related shapes $\{x_i\}_{i=1}^n$ for retrieval.
Specifically, we disentangle $f_b$ and $l_b$, where $f_b$ effectively extracts global-level structural features from the Scene Encoder $E_{S}$ through distillation to guide the INF in learning complex detailed object representations. Next, we constrain $l_b$ to multiple spherical layout attributes (such as position and scale) and use Layout Loss~\cite{sceneprior} to guide $l_b$ in capturing scene layout relationships from the Scene Encoder $E_{S}$, as illustrated on the left of Fig.~\ref{fig:field}.
Instead of using traditional 3D-CNNs, we implement the Scene Encoder $E_{S}$ with sparse CNNs in distillation, which robustly encode and compress the scene through an efficient hierarchical structure. 
The scene is compressed into a latent vector $z$, then mapped into the INF $f_b$ and layout $l_b$ using the Relationship Decoder $D_{RS}$. 
To facilitate sampling, we constrain the latent distribution $p(Z)$ to approximate a Gaussian distribution.

\subsubsection{Detailed Guidance.}\label{sec:render}
In addition to learning realistic multimodal relationships, ISS must further consider style consistency when capturing detailed object relationships. 
However, current methods, which rely on oversimplifying explicit representations, struggle to establish enough connections in detailed object relationships, leading to inconsistent object detailed styles.

To encourage style-consistent detailed object relationships, we utilize differentiable rendering to generate dense, detailed style guidance, enhancing style-detailed awareness of object components within the S-INF, as shown on the right side of Fig.~\ref{fig:field}.
Specifically, we randomly initialize a camera pose on the spherical space. Then, we render each object within the scene individually (related shapes or ground truths) to obtain their normals $\hat{\sigma}_{n}$ or ${\sigma}_{n}$ and masks $\hat{\sigma}_{m}$ or ${\sigma}_{m}$.
Note that the normals $\hat{\sigma}_{n}$ or ${\sigma}_{n}$ equip $\hat{\sigma}_{normal}$ or ${\sigma}_{normal}$ in Fig.~\ref{fig:method}, and the mask $\hat{\sigma}_{m} = \hat{\sigma}_{mask}$ and ${\sigma}_{m} = {\sigma}_{mask}$.
We render each object independently to reduce occlusion effects while preserving their positional information within the scene.

Through this rendering process, we implicitly acquire style information in detailed object relationships, while encouraging the S-INF to adaptively integrate detailed object relationships with scene layout relationships at all levels, creating realistic and style-consistent related shapes for retrieval.
We employ the Render Loss $L_{RD}$ as follows:
\begin{equation}
\begin{matrix}
L_{RD} = ||\hat{\sigma}_{n} - {\sigma}_{n}||_1 + \mathrm{MaskIOU}(\hat{\sigma}_{m},{\sigma}_{m}), 
\end{matrix}
\end{equation}
where the $\mathrm{MaskIoU}(\cdot, \cdot)$ is the Mask IoU loss.

\subsection{Training and Inference}

\subsubsection{Training.}
In our implementation, the Scene Encoder $E_S$ compresses the scene $X$ into a latent vector $z$.
Then, the Relationship Decoder $D_{RS}$ decomposes the latent vector $z$ into the layout $l_b$ and the INF $f_b$, where the layout $l_b$ is implemented through the initial transformation of template spheres for all objects. 
To fine-tune the layout $l_b$ into detailed related shapes $\{x_i\}_{i=1}^n$, the Instance Decoder $D_{IS}$ projects the layout $l_b$ into the INF $f_b$ and obtains refined deformations at each point, applying a secondary transformation to the initially transformed template spheres to generate the related shapes $\{x_i\}_{i=1}^n$.
Following previous work~\cite{sceneprior}, we then use Chamfer distance for shape retrieval to obtain the final scene $\hat{X}$.
We train our network end-to-end with the following loss:
\begin{equation}
\mathcal{L} = \alpha\mathcal{L}_\mathrm{KL} + \mathcal{L}_\mathrm{RD} + \mathcal{L}_\mathrm{LO},
\end{equation}
where $\mathcal{L}_\mathrm{LO}$ represents the Layout Loss, and $\mathcal{L}_\mathrm{KL}$ is the Kullback-Leibler Loss, with $\alpha = 1\times 10^{-4}$.

\subsubsection{Inference.}
During the inference stage, we directly sample the latent vector $z$ from the latent space $p(Z)$, then follow the training process and decode the latent vector $z$ to the related shapes $\{x_i\}_{i=1}^n$, then finally use the Chamfer distance to retrieval shapes and replaced with the final object meshes.

\begin{figure*}[!ht]
\centering
\includegraphics[width = 0.9\textwidth]{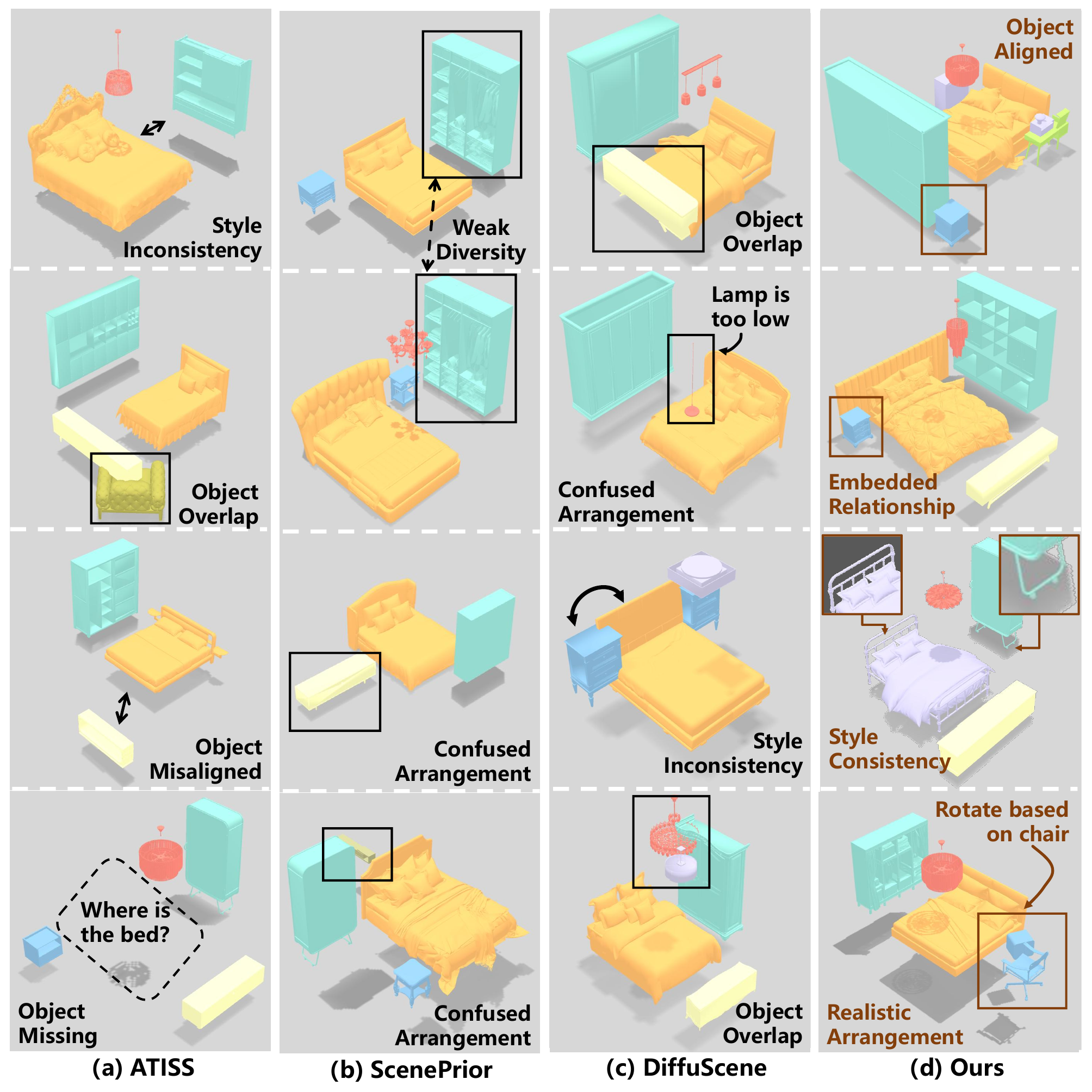}
\caption{Qualitative comparisons of bedroom with different baselines on ISS.}\label{fig:result}
\end{figure*}

\section{Experiments}

\subsection{Experiment Setup}\label{sec:baseline}

\subsubsection{Datasets.}To verify the performance in different scene types, we choose three scene types: Bedroom, Living Room, and Dining Room from the 3D-FRONT~\cite{3dront}.

\subsubsection{Baselines.}Apart from previous state-of-the-art methods such as Sync2Gen (SG), ATISS (AT), ScenePrior (SP), DiffuScene (DS), EchoScene (ES), and InstructScene (IS), we also provide variants methods for a full comparison, designed based on the mentioned baselines. 
We design the ScenePrior-NN (NN) to utilize a MLP as $D_{RS}$, passing the code through three fully connected layers, projecting it onto a concatenation vector, and dividing it into equal lengths.
On the other hand, the ScenePrior-Tr (Tr) employs a Transformer~\cite{transformer} as $D_{RS}$ to parallel generate permutation-invariant layout representations.
For a fair comparison, we optimize all the only-2D-supervision baselines using 3D IoU as layout loss, which has been admitted with a more stable performance in training~\cite{sceneprior}.

\subsubsection{Implementation.} We provided all baselines with groundtruth meshes, categories, positions, render masks, and normal images in training.
Given variations in scene sizes, we normalize all scene sizes to the range of $[-0.5,0.5]$. 
Optimization is carried out using the AdamW~\cite{adamw} optimizer with a batch size of 4 and a learning rate of 
$1\times10^{-4}$. 
All experiments are conducted on one RTX3090 GPU.

\subsubsection{Evaluation Metrics.} 
Similar to~\cite{atiss}, we utilize the Fréchet Inception Distance (FID), category KL divergence (KL), and scene classification accuracy (SCA) for quantitative comparisons.
We normalize all the scenes first, then select a top-down projection to render $1024^2$ images.
Rather than generating category-color maps, we generate normal images for more detailed structure comparison.
Also, to compare the diversity (Div) of different methods, we calculated the average $\mathcal{L}_2$ distance between all pairs of scene-rendered images, which is computed as follows: $\frac{1}{N(N-1)}\sum_{i=1}^N\sum_{j=1}^N||\hat{\sigma}_n-{\sigma}_n||_2$. 
Note that lower KL and FID values indicate better performance, higher Div means higher diversity and the best SCA score is 0.5.

\begin{table*}[!ht]
\centering
\tabcolsep=1.7mm
\begin{tabular}{cccccccccccccccc}
\toprule[1.5pt]
          &    & \multicolumn{4}{c}{Bedroom}  & & \multicolumn{4}{c}{Living Room}  & & \multicolumn{4}{c}{Dining Room}\\ \cmidrule{3-6} \cmidrule{8-11} \cmidrule{13-16} 
 Met.  &  Re.  & FID${}^\downarrow$  & KL${}^\downarrow$  & SCA${}^\sim $ & Div${}^\uparrow$ & & FID${}^\downarrow$  & KL${}^\downarrow$  & SCA${}^\sim $ & Div${}^\uparrow$ & & FID${}^\downarrow$  & KL${}^\downarrow$  & SCA${}^\sim $ & Div${}^\uparrow$\\ \midrule
NN     & $\times$     & 161.39  & 0.2521 & 0.8832  & 0.7284 &  & 188.25   & 0.1455   & 0.8565  & 0.7354 &  & 203.34 &0.4252 & 0.9163 & 0.7346\\
Tr      & $\times$   & 156.42  & 0.2082 & 0.8742  & 0.7271 & & 156.48   & 0.1339   & 0.8583  & 0.7414   &  & 162.82 &0.2528 &0.8821&0.7456\\
SP        & $\times$  & 164.70  & 0.2165 & 0.8596 & 0.7338 &  & 172.31   & 0.1357   & 0.8205  & 0.7390  && 168.85 &0.2640& 0.8812& 0.7441\\ \midrule
Ours    & $\times$      & \bf 131.47  & \bf 0.1985 & \bf 0.8393  & \bf 0.7379 &  & \bf 155.47   & \bf 0.1334   & \bf 0.8034 & \bf 0.7431 && \bf 157.93 &\bf 0.2342 & \bf 0.8476 & \bf 0.7465 \\ \midrule[1.5pt]
SG      & $\surd$       & 119.68  & 0.2403 & 0.8128  & 0.7083 &  & 179.32   & 0.2190   & 0.8669  & 0.7260 && 99.97 & 0.2866 & 0.7062 & 0.7146\\
AT       & $\surd$          & 118.38  & 0.2069 & 0.7632 & 0.7161 &  & 174.13   & 0.3024   & 0.8871  & 0.7320 && 131.16 & 0.2397 & 0.7802 & 0.7063\\
NN    & $\surd$     & 114.63  & 0.2521 & 0.7105 & 0.7072 &  & 81.34    & 0.1455   & 0.6455  & 0.7188 && 84.46&0.4252&0.6437&0.7174 \\
Tr    & $\surd$      & 117.02  & 0.2082 & 0.7865 & 0.7161 &  & 85.46    & 0.1339   & 0.6582  & 0.7106 && 97.55&0.2528&0.7020&0.7094 \\
SP     & $\surd$       & 110.74  & 0.2165 & 0.7994 & 0.7223 &  & 84.36    & 0.1357   & 0.6708  & 0.7269 && 131.20 & 0.2640 & 0.8792 & 0.7036 \\
ES   & $\surd$       & 107.27  & - & 0.6994  & 0.7259 &  & 109.30    & -   & 0.6792  & 0.7275 && 119.30 & - & 0.7500 & 0.7208 \\ 
DS    & $\surd$     & 84.40  & 0.2060 & 0.6319  & 0.7118 &  & 82.00    & 0.1771   & 0.6433  & 0.7319 && 86.28&0.2464 & 0.6533 & 0.7206 \\ 
IS   & $\surd$    & 89.68  & - & 0.6202  & 0.7231 &  & 80.15    & -   & 0.6435  & 0.7233 && 86.55 & - & 0.6706 & 0.7285 \\ 
\midrule
Ours  & $\surd$  & \bf 79.52   & \bf 0.1985 & \bf 0.6128 & \bf 0.7319 &  & \bf 78.98    & \bf 0.1334   & \bf 0.6371  & \bf 0.7322 && \bf 79.95 & \bf 0.2342 & \bf 0.6236 & \bf 0.7297 \\ \bottomrule[1.5pt]
\end{tabular}
\caption{We conducted a quantitative comparison of our method with state-of-the-art approaches on the 3D-FRONT dataset, where our method consistently demonstrated superior performance. Note that since EchoScene (ES) and InstructScene (IS) are class-conditional, their category-KL divergence is excluded, "Re" is retreival and "$\sim$" in SCA indicates that 0.5 is optimal.}
\label{tab:exp}
\end{table*}

\subsection{Experiment Results}

\subsubsection{Quantitative Comparisons.} Table~\ref{tab:exp} provides a quantitative comparison in 3D-FRONT. 
Without retrieval, the compared methods rely on over-simply explicit representations to model scenes, discarding the multimodal relationships within the scenes. As a result, they fail to generate sufficiently complex scene representations, reflected in poor FID and SCA performance after retrieval. In contrast, by decoupling and learning the multimodal relationships within scenes, S-INF leverages scene layout relationships to generate more realistic and diverse scenes, while utilizing detailed object relationships to ensure the stylistic consistency of related shapes. This makes the retrieval results more diverse and has a better FID, SCA, and diversity.

In addition, during retrieval, although existing methods like DiffuScene\cite{diffuscene} and InstructScene\cite{instructscene} achieve good FID and SCA performance, they overlook detailed object relationships in scenes and focus solely on modeling scene layout relationships. This results in a lack of diversity in their generated outcomes and leads to style-inconsistent results.
From the performance in all metrics, our S-INF has a more outstanding performance in both realistic modeling and style-consistency.

\subsubsection{Qualitative Comparisons.} We present the qualitative comparative results in Fig.~\ref{fig:result}. 
All methods employ Chamfer Distance to gauge the similarity between $x_i$ and CAD models in 3D-FUTURE~\cite{3dfuture}. 
Unlike previous baselines that handle objects independently, we fully consider the multimodal relationships in the scene, enabling S-INF can generate realistic and style-consistent results.

\begin{figure}[!ht]
\centering
\includegraphics[width = 0.445\textwidth]{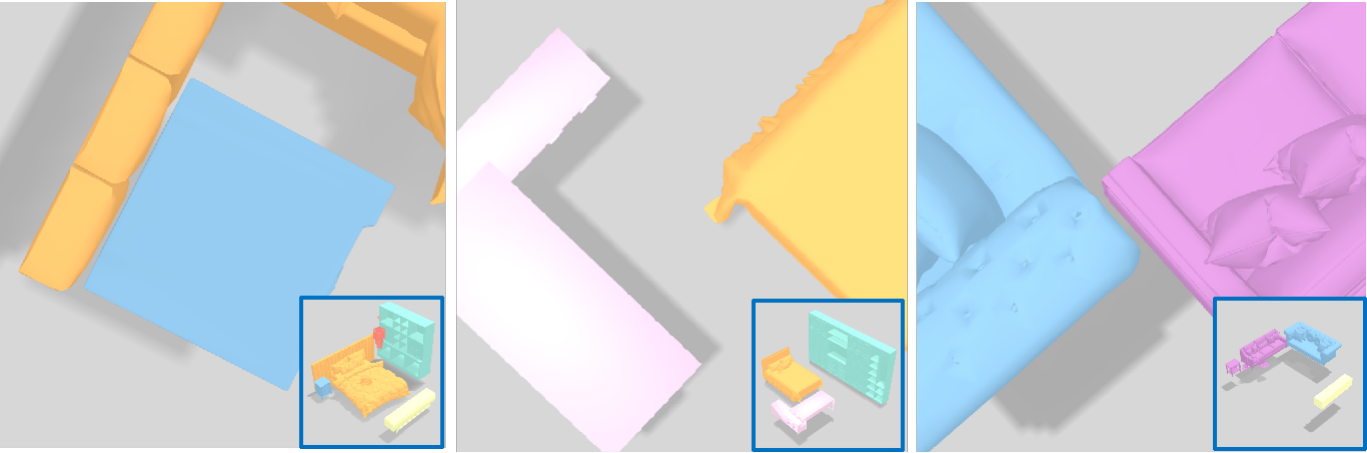}
\caption{Realistic ISS from multimodal relationships.}
\label{fig:pos}
\end{figure}

\begin{figure}[!ht]
\centering
\includegraphics[width = 0.445\textwidth]{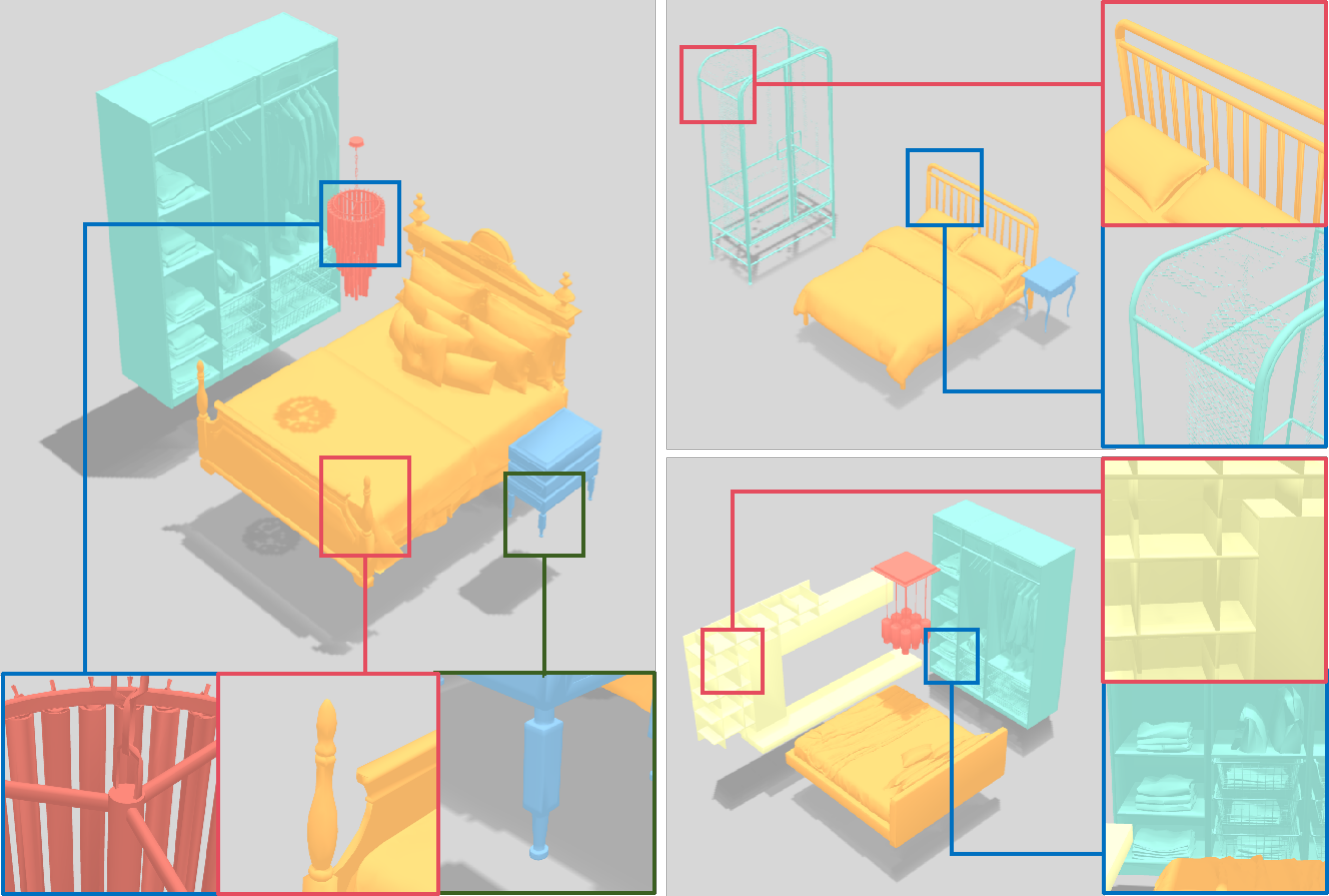}
\caption{Style consistancy ISS from detailed object relationships.}
\label{fig:style}
\end{figure}

We also visualize the S-INF generation capabilities in Fig.~\ref{fig:pos} and Fig.~\ref{fig:style}. 
The results show that the S-INF efficiently captures realistic multimodal relationships, highlighting that S-INF can generate tighter scenes without any overlap, misalignment, unaligned, or confused arrangement.
Fig.~\ref{fig:style} shows scenes with detailed object relationships. 
After retrieval, each scene presents objects unified by similar object details. 
This visualization highlights the ability of the S-INF to create realistic layout and uniform style-related detail scenes, ensuring realistic and consistent ISS.

\begin{table}[!ht]
\centering
\begin{tabular}{cccccc} \toprule[1.5pt]
Type & $D_{RS}$ & FID${}^\downarrow$  & KL${}^\downarrow$  & SCA${}^\sim $ & Div${}^\uparrow$\\ \midrule
&$l_b + f_b $& \bf 131.47 &\bf 0.1985 &\bf 0.8393& \bf 0.7309\\
Bed&$l_b$    & 156.42 & 0.2082 & 0.8742 &  0.7271\\
 &$f_b$   & 184.86 & 0.2408 & 0.8526& 0.7137\\
  \midrule
&$l_b + f_b $  & \bf 155.47 & \bf 0.1334 & \bf 0.8034& \bf 0.7431\\
Living&$l_b$& 156.48 & 0.1339 & 0.8583 & 0.7414 \\
 &$f_b$   & 176.85 & 0.5819 & 0.8690&  0.7193 \\\bottomrule[1.5pt]
\end{tabular}
\caption{Evaluation of the configurations of $D_{RS}$. Note that "$l_b + f_b $" denotes both $l_b$ and $f_b$ are both provided.}
\label{tab:exp1}
\end{table}

\subsubsection{Ablation Study.}
Table~\ref{tab:exp1} compares the effects of different configurations of the relationship decoder $D_{RS}$, where the “$l_b + f_b$” setting refers to the scene layout representation $l_b$ (which learns scene layout relationships) and INF $f_b$ (which learns detailed object relationships). Note that $l_b$ and $f_b$ imply that we ignored disentangling and modeling the INF and layout during training, but strived to maintain the parameter amount the same with the “$l_b + f_b$”. 

The results indicate that modeling the two relationships in a disentangled manner achieves the best performance; omitting either leads to a significant performance drop.

\section{Conclusion}
In this paper, we propose a novel 3D ISS method called Scene Implicit Neural Field (S-INF).
S-INF effectively captures multimodal relationships within scenes, enhancing the realistic and style-consistency of ISS.
It directly distillates more advantageous multimodal relationships from the entire scene, effectively capturing both scene layout relationships and detailed object relationships.
For generating realistic related shapes for retrieval, S-INF achieves realistic multimodal relationship learning by disentangling and modeling scene layout relationships into the layout and detailed object relationships info the INF.
For style-consistancy, differentiable rendering is employed to enrich style information across objects.
Extensive experiments on widely used benchmarks show that our method consistently achieves state-of-the-art performance in ISS tasks.

\section{Acknowledgements}
This research was partially funded by the National Natural Science Foundation of China (No. 82121003, and No. 62176047), the Shenzhen Fundamental Research Program (No. JCYJ20220530164812027). 

\bibliography{aaai25}

\end{document}